\documentclass{ws-procs11x85}
\usepackage{ws-procs-thm}           % comment this line when `amsthm / theorem / ntheorem` package is used
\usepackage[utf8]{inputenc} % allow utf-8 input
\usepackage[T1]{fontenc}    % use 8-bit T1 fonts
\usepackage{hyperref}       % hyperlinks
\usepackage{url}            % simple URL typesetting
\usepackage{booktabs}       % professional-quality tables
\usepackage{amsfonts}       % blackboard math symbols
\usepackage{nicefrac}       % compact symbols for 1/2, etc.
\usepackage{microtype}      % microtypography
\usepackage{xcolor}         % colors
\usepackage{graphicx}
\usepackage{amsmath}
\definecolor{raquel}{RGB}{102,204,0}

\begin{document}

\title{Multi-treatment Effect Estimation from Biomedical Data}

\author{Raquel Aoki$^\dag$, Yizhou Chen and Martin Ester}

\address{School of Computing Science, Simon Fraser University,\\
Vancouver, British Columbia, Canada\\
$^\dag$E-mail: raoki@sfu.ca}

\begin{abstract}

Several biomedical applications contain multiple treatments from which we want to estimate the causal effect on a given outcome. Most existing Causal Inference methods, however, focus on single treatments. In this work, we propose a neural network that adopts a multi-task learning approach to estimate the effect of multiple treatments. We validated M3E2 in three synthetic benchmark datasets that mimic biomedical datasets. Our analysis showed that our method makes more accurate estimations than existing baselines. 
\end{abstract}

\keywords{Causal Inference, Multiple-treatments, biomedical data}

% required, do-not-remove
\copyrightinfo{\copyright\ 2022 The Authors. Open Access chapter published by World Scientific Publishing Company and distributed under the terms of the Creative Commons Attribution Non-Commercial (CC BY-NC) 4.0 License.}

% \begin{center}
% {\tablefont
% \begin{tabular}{ll}
% \toprule
% Environment & Heading\\\colrule
% \verb|algorithm| & Algorithm\\
% \verb|answer| & Answer\\
% \verb|assertion| & Assertion\\
% \verb|assumption| & Assumption\\
% \verb|case| & Case\\
% \verb|claim| & Claim\\
% \verb|comment| & Comment\\
% \verb|condition| & Condition\\
% \verb|conjecture| & Conjecture\\
% \verb|convention| & Convention\\
% \verb|corollary| & Corollary\\
% \verb|criterion| & Criterion\\
% \verb|definition| & Definition\\
% \verb|example| & Example\\
% \verb|lemma| & Lemma\\
% \verb|notation| & Notation\\
% \verb|note| & Note\\
% \verb|observation| & Observation\\
% \verb|problem| & Problem\\
% \verb|proposition| & Proposition\\
% \verb|question| & Question\\
% \verb|remark| & Remark\\
% \verb|solution| & Solution\\
% \verb|step| & Step\\
% \verb|summary| & Summary\\
% \verb|theorem| & Theorem \\\botrule
% \end{tabular}}\label{aba:theo}
% \end{center}

\section{Introduction}\label{introduction}

Consider the following setting: an exploratory study on hearing loss as an Adverse Drug Reaction (ADR) in children under cancer treatment with the drug Cisplatin\cite{drogemoller2019pharmacogenomics}. While Cisplatin is one of the most effective chemotherapeutic agents for children, reports have also demonstrated that 75-100\% of infant patients have hearing loss. Note that patients often receive a drug cocktail, and while a single drug might not lead to ADR, ADR is observed when we have a combination of these drugs. Previous studies\cite{drogemoller2019pharmacogenomics} pointed out that hearing loss is the result of a combination of factors, such as the patient's age, genetic predisposition, dosage, and exposure to several drugs (more drugs, more heavy metals accumulation in the body, higher the chances of hearing loss). The study's data are the patient's clinical information (low-dimensional), genetic information (high-dimensional), the drugs given to the patient, and the observed ADR. %Learning the treatment effect of interventions on the outcome of interest can help improve therapy recommendations by reducing negative ADRs.

In Causal Inference notation, the covariates $X$ are the patients' clinical information and genetic information; the outcome of interest $Y$ is the ADR, and each drug is a binary treatment ($\mathcal{T} = [T_0, T_1,..., T_K]$, where $T_k=1$ records that the $k$-th drug was given). Understanding and learning the causal effect of each treatment on the outcome can be used to support doctors in recommending more precise treatments, minimizing ADRs in this example, or maximizing the drug response in other cases. Note that existing treatment effect estimators designed for individual binary treatments could be adopted: For each drug $k\in\{0,..., K\}$, we fit an estimator using all the other drugs as covariates. However, such an approach assumes the estimator would perform covariate adjustment correctly - and here is where we argue that an estimator that considers the multiple treatments together could be a better alternative for biomedical data.

Recent advances in Machine Learning(ML) are now widely being used to improve Causal Inference methodologies. One example is how ML can improve the covariate adjustment of applications with high-dimensional datasets. Such improvements fit perfectly with the precision medicine vision of developing diagnosis, prognosis, and treatment techniques that consider the individual, often high-dimensional data. Most machine learning methods solve only a single task, i.e. they predict a single target variable. Multi-task learning (MTL) methods  \cite{ruder2017overview}, on the other hand, optimize a model to simultaneously solve multiple tasks (or, in our context, treatments). The main argument in favor of MTL is that single-task learning may fail to capture the synergy of multiple treatments, e.g., an additive effect or a genetic predisposition to a certain combination of treatments, but not to individual treatment. Currently, there are only a few methods capable of estimating the causal effect of multiple treatments. Hi-CI \cite{sharma2020hi} considers and models multiple treatments but assumes that only one is assigned to a unit at any given time. The Deconfounder Algorithm (DA) \cite{wang2019blessings}, a probabilistic graphical model, works with multiple treatments but has received some recent criticism regarding its assumptions \cite{d2019multi}.

%In this paper, we propose Multi-gate Mixture-of-experts for Multi-treatment Effect Estimation (M3E2): an MTL-based neural network for estimating the causal effect of multiple treatments. 

\textbf{Contributions}: The main contributions of this paper are as follows: 
\begin{itemize}
    \item We propose the Multi-gate Mixture-of-experts for Multi-treatment Effect Estimation (M3E2), a method to estimate the multi-treatment effect. %with an MTL neural network architecture.
    \item We validate M3E2 in three synthetic datasets that mimic biomedical applications. We also compare our method with three existing baselines. 
    \item We create the repository \url{github.com/raquelaoki/M3E2} with an implementation of our methods, baselines, and datasets. We also share all the configuration files for reproducibility of our results, with hyperparameters and seeds adopted. 
\end{itemize}

\section{Related Work}
This work combines the estimation of treatment effects and multi-task learning (MTL).

\textbf{Estimating Treatment Effects}: BART \cite{hill2011bayesian}, Causal Forests \cite{wager2018estimation}, CEVAE \cite{louizos2017causal}, and Dragonnet \cite{shi2019adapting}, have explored the estimation of a single treatment effect, using Bayesian Random Forests, Random Forests, VAEs, and neural networks (NN) respectively. The inverse propensity weighting-based methods\cite{hernan2006estimating}, meta-learners\cite{curth2021nonparametric} also focused on binary single-treatments. The Deconfounder Algorithm \cite{wang2019blessings}, Hi-CI \cite{sharma2020hi}, approaches based on the propensity score \cite{lechner2001identification, lopez2017estimation}, and others \cite{miao2021identifying, tanimoto2021regret, qian2021estimating} aim to estimate multi-treatment effect.  However, many of these methods assume that only one treatment is applied to any given unit or consider all the combinatorial interventions, which is infeasible for larger numbers of treatments. Note that several works assume robustness to missing confounders \cite{louizos2017causal, wang2019blessings, miao2021identifying, mastouri2021proximal}. Their robustness is often built on the assumption that extra information is known, such as a known number of hidden confounders or replacing unobserved confounders with proxies. There are, however, several concerns regarding some of these methods \cite{d2019multi, rissanen2021critical}. Our proposed method focuses on multiple treatment effect estimation through an outcome model in a multi-task learning neural network architecture and ignorability. By considering all treatments simultaneously, our proposed architecture can learn a better representation of input data and perform a  better covariate adjustment than existing baselines.

\textbf{Multi-task learning (MTL)}: 
MTL neural network (NN) architectures aim to optimize a single model for two or more tasks simultaneously. Hard-parameter sharing NN\cite{caruana1993multitask} is one of the MTL pillars. Such architecture is composed of a set of layers shared among all tasks and a set of task-specific layers on the top. From the MTL perspective, the Dragonnet \cite{shi2019adapting} has a hard-parameter sharing architecture. Building upon the hard-parameter sharing architectures, the Multi-gate Mixture-of-Experts (MMoE)\cite{ma2018modeling} architecture, where each expert can be seen as a hard-parameter sharing NN, and all the experts are combined through a gate function, which is also trainable. The core idea of such an approach is to improve the model's generalization; plus, it allows experts to specialize in one of the tasks. To put into perspective, an MMoE is to hard-parameter sharing NN what a Random Forest Model is to a Decision Tree. Our proposed method M3E2 uses a MMoE \cite{ma2018modeling} as a component. Our work expands the MMoE architecture to satisfy causal inference assumptions and estimate the multi-treatment effect.

%\subsection{Multi-task learning}

\section{MMoE for Multi-treatment Effect Estimation} \label{method}

This section describes our proposed method, M3E2. Its multi-task learning architecture simultaneously predicts the outcome and the propensity scores for each treatment.

\begin{figure}[h]
    \centering
     \includegraphics[scale=0.45]{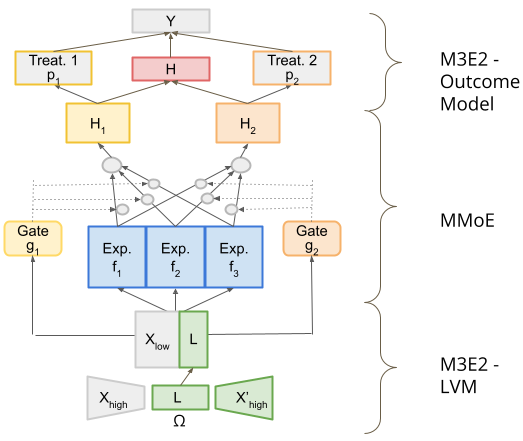}
         \caption{M3E2 training architecture, for $K=2$ (two treatments), and $3$ experts. It receives as input the covariates $X=[X_{low},X_{high}]$, and predicts the treatment assignment $\mathcal{T} = \{T_1,...,T_K\}$ and the outcome $Y$. The LVM model $\Omega$ learns a latent representation $L$ of the high-dimensional covariates $X_{high}$. The gates $g_k$, experts $f_e, \forall e\in\{1,2,3\}$, and task-specific layers $H_1$ and $H_2$ learn a representation $H$ of the input data, and $H$ is used to predict the propensity scores $p_1$ and $p_2$ and the outcome $Y$.}
     \label{fig1_a}    
\end{figure}

When working with observational studies, one must always describe how the confounders are addressed. Some works assume no unobserved confounders \cite{shalit2017estimating, shi2019adapting,glynn2010introduction, hill2011bayesian}, others try to reduce the bias through latent variables \cite{louizos2017causal, wang2019blessings, mastouri2021proximal}; while others question if the latent variables are solving the problem at all \cite{d2019multi, rissanen2021critical}. While exploring alternatives to the  ignorability assumption is an interesting research direction, the main focus of this work is the estimation of effect of multiple treatments. Hence, in our work, we assume no unobserved confounders.

Figure \ref{fig1_a} illustrates the proposed neural network architecture, with a MMoE \cite{ma2018modeling}, and a Latent Variable Model (LVM) as subcomponents. This architecture predicts $K+1$ tasks: the outcome $Y$ and $K$ propensity scores $p_k$. The propensity scores estimate the probability of a treatment being assigned given the covariates ($P(T_k=1|X)$), and it is important to guarantee the identifiability of the causal effects (Theorem \ref{theorem:suffiency}). The LVM contributes to the model by efficiently combining low and high-dimensional covariates (section \ref{autoencoder}). The MMoE is an MTL architecture adopted to handle multiple tasks. It contains a combination of experts, gates, and task-specific layers (section \ref{experts}).

One of the strengths of M3E2 is its capacity to estimate the combined effect of a large number of treatments: the M3E2 network only grows linearly with the number of treatments, handling all potential combinations, something that other multi-treatment methods typically struggle to accomplish. Furthermore, the proposed architecture of M3E2 extends the MMoE architecture by incorporating causal inference assumptions through suitable regularizers and adding the outcome model to estimate the treatment effects.

\textbf{Notation:} We define low-dimensional covariates as $X_{low}$ and high-dimensional covariates as $X_{high}$. An example of the first is clinical variables and, from the latter, genomics information. The split of covariates into low-dimensional and high-dimensional will be explained in Section \ref{autoencoder}. We define the covariates concatenation as $X = [X_{low},X_{high}]$. The continuous outcome is $Y$, and $K$ represents the number of treatments. 
$\mathcal{T} = \{T_0=t_0, T_1=t_1,...,T_K=t_K\}$, where $\mathcal{T}$ could e.g. be the \textit{drug cocktail} taken by a patient.

\subsection{Assumptions}\label{assumption}

\begin{assumption}\label{assumption:sutva}
    Stable Unit Treatment Value Assumption (SUTVA) \cite{rubin1980randomization}: the response of a particular unit depends only on the treatment(s) assigned, not the treatments of other units. 
\end{assumption}

\begin{assumption}\label{assumption:multiple}
Common Confounders and conditional independence \cite{ranganath2018multiple}: Treatments share confounders.  Given the shared confounders, the treatments are independent. 
$$T_i \perp T_j | X, \forall i, j \in \{0,...,K\}, i\neq j$$
\end{assumption}

\begin{assumption}\label{assumption:ignorability}
    Ignorability - the potential outcome is independent of the treatments given the covariates.
\end{assumption}

\begin{theorem}\label{theorem:suffiency}   
Sufficiency of Propensity Score\cite{rosenbaum1983central,shi2019adapting}: If the average treatment effect is identifiable from observational data by adjusting for $X$, i.e., $ATE = \mathbb{E}_X[\mathbb{E}_Y[Y|X,T = 1] - \mathbb{E}_Y[Y|X, T = 0]]$, then adjusting for the propensity score also suffices: 

$ATE = \mathbb{E}_X[\mathbb{E}_Y[Y|h(X),T = 1] - \mathbb{E}_Y[Y|h(X), T = 0]]$
\end{theorem}

First, we consider applications with a continuous outcome, binary or continuous treatments, and a set of covariates.
Assumption \ref{assumption:sutva} (SUTVA) is standard in Causal Inference. According to SUTVA, the samples are independent and do not interfere with each other. Assumptions \ref{assumption:multiple} and \ref{assumption:ignorability} are related to the identifiability of the treatment effect. Assumption \ref{assumption:multiple} assumes no links (dependencies) between the treatments given the covariates, and Assumption \ref{assumption:ignorability} assures all back-door paths can be blocked by conditioning on the observed covariates $X$ - guaranteeing the identifiability of the treatment effect \cite{pearl1995causal}. Assumption \ref{assumption:multiple} is also related to multi-task learning (MTL). The ideal use of MTL is when tasks (in our case, treatments) are somehow related. In that case, it is reasonable to assume they also share confounders. The Theorem \ref{theorem:suffiency} is presented here as originally proposed, so for the proofs and demonstrations, please check the original publications \cite{rosenbaum1983central, imbens2000role}. According to Theorem \ref{theorem:suffiency}, it suffices to adjust only the information in $X$ that is relevant for predicting the treatment $T_k$, which is the output of $H_k(X_{L1})$. For multiple treatments, the generalization goes as follows \cite{imbens2000role}: 

$ATE = E[E[Y|H(X_{L1}), T_1 = t_1,...,T_K = t_K] - E[Y| H(X_{L1}),T_1 = 1-t_1,...,T_K = t_K]]$

Under these assumptions and theorem, the identifiability comes from the Propensity Score's Sufficiency and the following causal structure: $\mathcal{T}\rightarrow Y$, $X\rightarrow \mathcal{T}$, $X\rightarrow Y$.

\subsection{Latent Variable Model (LVM)}\label{autoencoder}

M3E2 can handle different data types by dividing the input covariates $X$ into two groups, $X_{low}$ and $X_{high}$. While the Latent Variable Model (LVM) handles the covariates in $X_{high}$, the $X_{low}$ covariates are fed directly to the experts. The split of the covariates $X$ into $X_{low}$ and $X_{high}$ is defined by the user. Ideally, $X_{high}$ contains high-dimensional covariates, such as gene expression, single-cell data, or image data; and $X_{low}$ contains low-dimensional data, such as clinical variables. Note that, in applications with only one data type, both $X_{low} = \emptyset$ and $X_{high} = X$, and $X_{low} = X$ and $X_{high} = \emptyset$ are acceptable splits.

In applications where $X_{high}\neq \emptyset$, M3E2 uses a LVM to reduce the dimensionality of the covariates in $X_{high}$. Note that, while there are similarities with other works that adopt proxies to handle unobserved confounders, our LVM component is responsible only for reducing the dimensionality of $X_{high}$. As described in Section \ref{assumption}, our work assumes strong ignorability, a setting with no unobserved confounders. Under strong ignorability, however, we can still have confounding within the observed data. The LVM component, along with the experts, is responsible for extracting a meaningful representation of the input data. These features are used in the covariate adjustment $E[Y | X, T_0, ..., T_k]$, which should close the back-doors and make the treatment effect identifiable. To learn a meaningful representation of $X$ in applications with a mix of high-dimensional and low-dimensional covariates, it was important to find an approach that is capable of combining these different types of covariates. Without the LVM component, the experts could give a disproportional weight to $X_{high}$ covariates, as they would be the majority in $X$, and even ignore relevant information in $X_{low}$.

In our experiments, M3E2 adopts an autoencoder with two linear encoder layers and two linear decoder layers. Note, however, that one is free to choose a different architecture or factor model to extract a latent representation of $X_{high}$. Consider an application with $n$ samples, $c_2$ columns in $X_{high}$, $c_L$ as the latent variables size, and the input data $X_{high}$ as a matrix $n\times c_2$. The function $\omega_{enc}(X_{high})$ returns $L_{(n\times c_L)}$, a representation of $X_{high}$ in a lower dimension. Finally, $\omega_{dec}(X_{high})$ returns the reconstructed data $X'_{high}$, back on $ n \times c_2$ space. %The autoencoder loss is the mean squared error between the input $X_{high}$ and the reconstructed input $X'_{high}$.

\subsection{MMoE Architecture}\label{experts}

In Machine Learning, it is common for a set of shared layers to predict multiple tasks. These architectures are called hard-parameter sharing neural networks. A multi-gate mixture-of-expert (MMoE)\cite{ma2018modeling} architecture contains several experts, where each expert can be seen as a hard-parameter sharing neural network. It was shown that MMoE architectures generalize better \cite{ma2018modeling}, especially in biological applications \cite{aoki2022heterogeneous}. 

The user defines the number of experts $E$ and the $f_e$ architecture. In the context of multiple treatment effect estimation, the tasks are the propensity score and the outcome $Y$ prediction. The experts' input data is $X_{L1} = [\Omega_{enc}(X_{high}),X_{low}] = [L, X_{low}]$. The ideal number of experts depends on the tasks. Homogeneous tasks might not benefit from many experts and might overfit if the number of experts is too large. Conversely, heterogeneous tasks tend to benefit from a larger number of experts. Note that the definition of homogeneous and heterogeneous tasks is subjective. Here, we define applications whose tasks adopt the same loss as homogeneous tasks. An example would be an application with only classification tasks. On the other hand, heterogeneous task applications contain classification, regression, multi-label, and other potential tasks in the MTL model. The gates control the contribution of each expert to each task. There is a gate $g_k$ per treatment defined as: $g_k(X_{L1}) = softmax(W_K \times X_{L1}), \forall k\in {1,...,K}$, where $W_k\in R^{E\times d}$ is a trainable matrix of weights, $E$ is the number of experts defined by the user, and $d$ is the number of columns in $X_{L1}$. Finally, note that the gates can be seen as an attention\cite{vaswani2017attention} mechanism, learning which experts are more relevant for each task.

\subsection{Task-specific Layers}\label{treatspecific}

The task-specific layers are responsible for predicting the propensity score $p_k$ and the outcome of interest $\hat{Y}$. Each treatment task-specific layer receives as input a weighted average of the experts, where the weights come from the gates associated with that given task. This relationship is formally defined as:

$H_k = h_k(\sum_{e=1}^Eg_k(X_{L1})f_e(X_{L1})), \forall k\in \{1,...,K\}$

In the training phase (Figure \ref{fig1_a}), the treatment assignment is predicted with the propensity score $p_k$, estimated as $p_k = P(T_k = t|H_k)$ (for discrete treatments) or $p_k = P(T_k \leq t|H)$ (for continuous treatments using the conditional density $f_{T|X}(t,x)$ \cite{hirano2004propensity, nie2021varying}). To estimate the treatment assignment of $T_k$ we only use $H_k,\forall k\in \{1,...,K\}$. For binary treatments, a softmax activation function will outputs, for each sample, the probability of $P(T_k=1|H_k)$ and $P(T_k=0|H_k)$. These predictions are used to calculate the loss of the neural network, as described in Section \ref{loss}. The propensity score losses are used to drive $H_k$ to be sufficient (Theorem \ref{theorem:suffiency} - Section \ref{assumption}). Note that $h_k$ can be a combination of one or more layers.

Finally, a layer with trainable weights $\Phi$ is used to predict the outcome. Consider the input data of this layer as $X_{TH} = [T_1, ...,  T_K, H]$, where $T_1,...,T_K$ are the observed treatment assignments, $H = \frac{\sum_{k=1}^K H_k}{K}$, and $c_{TH}$ is the number of columns. The trainable weights layer $\Phi = [\tau_1, …, \tau_k, …, \tau_{c_{TH}}]$ estimates the final outcome as $Y = \Phi \times X _{TH}$. In our context of treatment effect estimation, $\tau_k$ is the treatment effect of the treatment $k$. The $\Phi$ works as an outcome model and each weight associated with a $T_i, \forall i\in\{0,..., K\}$ represents an $ATE_i$.

Our approach targets additive effect, which is fairly common in biomedical applications\cite{zheng2021copula}. Consider, for example, the ADR study on patients under cancer therapy described in Section \ref{introduction}. Many of these drugs contain heavy metals, and their accumulation can result in adverse drug reactions. Non-linear effects\footnote{Note that the linearity only applies to the last layer $\Phi$, not to the autoencoder or the experts.} are an interesting extension left for future work.

\subsection{Loss function}\label{loss}

M3E2's loss function is composed of: 
\begin{enumerate}

    \item Root mean square error loss $\ell_y(Y,\hat{Y}) = RMSE(Y,\hat{Y})$ for continuous outcomes and binary cross-entropy $\ell_y(Y,\hat{Y}) = BCE(Y,\hat{Y})$ for binary outcomes. 
    
    \item Similar to the outcome loss functions, we adopt $\ell_{p_k}(T,T') = RMSE(T_k,\hat{T}_k)$ or/and $\ell_{p_k}(T,T') = BCE(T_k,\hat{T}_k)$ as the propensity score losses, $\forall k\in\{0,...,K\}$.

    \item $\ell_A(X_{high},X'_{high}) =RMSE(X_{high}, X'_{high})$ is the autoencoder loss function. 
    
    \item $\frac{1}{2n}\sum_ww^2$ as the $L_2$ regularization.

\end{enumerate}

As a reminder, while our architecture minimizes the propensity score and the outcome losses, our main target is to obtain estimates of the treatment effects. The treatment effects are a co-product of this model, i.e., the weights associated with the treatments in the trainable layer $\Phi$ (See Section \ref{treatspecific}). The model also learns weights in $\Phi$ associated with the $H$; however, these are not considered treatment effects. The total loss is $\mathcal{L} = \alpha\ell_y + \beta\sum_k^K\ell_{p_k} + \gamma\ell_A + \frac{\lambda}{2n}\sum_w w^2$, where $\alpha$, $\beta$ and $\gamma$ are weights. There are two possible ways to define these weights: to adopt them as a hyper-parameter or to adopt an MTL task balancing approach. Modifying both $\ell_{g_k}$ and $\ell_y$ to other loss functions is also straightforward.

\section{Experiments}\label{experiments}

In causal inference, the lack of ground truth for real-world applications poses a challenge to its evaluation. Therefore, we adopt three synthetic datasets that have known treatment effects. These synthetic datasets mimic existing biomedical datasets:

\begin{itemize}

\item Genome-Wide Association Study (GWAS)\cite{song2015testing, wang2019blessings, aoki2020parkca}: Semi-synthetic sparse dataset with 1000 covariates, 3-10 binary treatments, and continuous outcome. In this dataset, the covariates and treatments are single-nucleotide polymorphisms (SNPs), and the outcome represents a clinical trait. The simulation starts by removing highly correlated SNPs with linkage disequilibrium from the 1000 Genome Project (TGP)\cite{consortium2015global}. Then, a PCA extracts $c=5$ components from TGP, creating the genetic representation matrix $\Gamma_{v,c}$. The patients' representation matrix is generated as $\Pi_{n, c} \sim 0.9 \times Uniform (0, 0.5)$, where $n$ is the number of desire samples. The covariates are simulated as $X_{n, v} \sim Binomial(1, \Pi_{n, c} \times \Gamma_{v,c}^T)$.  The set $\mathcal{K}$ contains the index of $K$ columns randomly picked to be treatments. The effect of each covariate is defined as $\tau_i\sim Normal(0, 0.5)\forall i\in\mathcal{K}$(causal effect), else, $\tau_i=0$ (non-causal effect). Three groups were extracted using k-means($X$) to add confounding. Each group $l\in\{1,2,3\}$ has an intercept value $\lambda_l$ and noise distribution $\epsilon~\sim Normal(0,\sigma_l)$, $\sigma_l\sim InvGamma(3,1)$. The outcome is calculated as $Y = \sum_v\tau_vX_{n, v}+\lambda_{l_n}+\epsilon$. %(For more details, please refer to related works\cite{song2015testing, wang2019blessings}).

\item Copula \cite{zheng2021copula}: This recently proposed dataset also mimics a Genome-Wide Association Study. The Copula, unlike the GWAS dataset, features a fully synthetic dataset. We adopted the setting with four treatments and non-linear outcomes. The covariates are generated as $X_{n,v}\sim Normal(0,\sigma)$, where $n$ is the sample size and $v$ the number of covariates. The treatments are simulated as $T_{n,l} = PCA_1(X_{n,v})+\epsilon_t, \forall l\in\{1,2,3,4\}$, $\epsilon_t\sim Normal(0,\sigma_t)$, and $Y = 3\times T_1-T_2+T_3I_{T_3>0}+0.7\times T_3I_{T_3\leq 0}-0.06\times T_4-4\times T_1^2 + 2.8\times \sum_vX_{n,v} + \epsilon_y$, $\epsilon_y\sim Normal(0, \sigma_y)$. The causal effects are $\tau=[1, 0.25, -0.2, 0.1]$.

\item IHDP \cite{hill2011bayesian, louizos2017causal,shi2019adapting}: the  Infant Health and Development Program (IHDP) is a traditional benchmark for single binary treatments. It is supposed to mimic a study on infant development. In that study, the treatment was assigned ($T=1$) if the child had special care/home visits from a trained provider. The outcome $Y$ is cognitive test scores, and the goal is to measure the causal effect of the home visits. This benchmark contains ten replications of such a study, with 24 covariates and a continuous outcome. We adopt this dataset to compare our proposed method with some of the single-treatment baselines that have been previously evaluated on the IHDP benchmark datasets. \footnote{Implementation available at \url{github.com/AMLab-Amsterdam/}}

\end{itemize}

Due to the synthetic nature of the datasets adopted \footnote{Implementation available at \url{github.com/raquelaoki/CompBioAndSimulated_Datasets}}, we can calculate the mean absolute error (MAE) between the estimated treatment effect and the true treatment effect. Defining $\tau_k$ as the true treatment effect of $T_k$, and $\hat{\tau}_k$ as its estimated value by one of the methods. As we have multiple treatment effects, we report their average error $\frac{\sum_{k=0}^K|\tau_k-\hat{\tau}_k|}{K}$, where $K$ is the total number of treatments. We repeat each combination of $(data \times model \times setting)$ $B=20$ times, and in our plots, we show the MAE calculated over all these runs:

\begin{equation}
    MAE = \sum_{b=0}^B\left( \frac{\sum_{k=0}^K|\tau_k-\hat{\tau}_k|}{K}\right)\frac{1}{B}
\end{equation}

A good estimator has estimates close to the true treatment effect values; therefore, \textit{low MAE values are desirable}. We adopt an experimental setting similar to the multi-task learning settings \cite{ma2018modeling}, where the proposed multi-task learning method is compared with other multi-task learning methods and single-task learning models. Among our baselines, the DA\cite{wang2019blessings} is the only method that can estimate the effect of multiple treatments with one model. The CEVAE \cite{louizos2017causal} and Dragonnet \cite{shi2019adapting} are single-treatment methods. We used the author's implementation of the baselines when available. For single-treatment baselines, the multiple treatment effects were estimated as follows: to estimate $\tau_1$, the baseline methods receive as input $T_1$ as the treatment assignment, and the columns $T_0, T_2, ..., T_K$ are added to $X_{low}$. We follow this setup for all $K$ treatments. We also performed experiments with BART. However, since CEVAE and Dragonnet achieved better performance results in the recent publications \cite{louizos2017causal, shi2019adapting}, and BART performed poorly on the GWAS and Copula datasets, we decided not to discuss BART in the experimental section.

\subsection{Overall Performance}

\begin{figure}[!h]
%    \centering
    \centerline{\includegraphics[width=0.7\textwidth,scale=0.45]{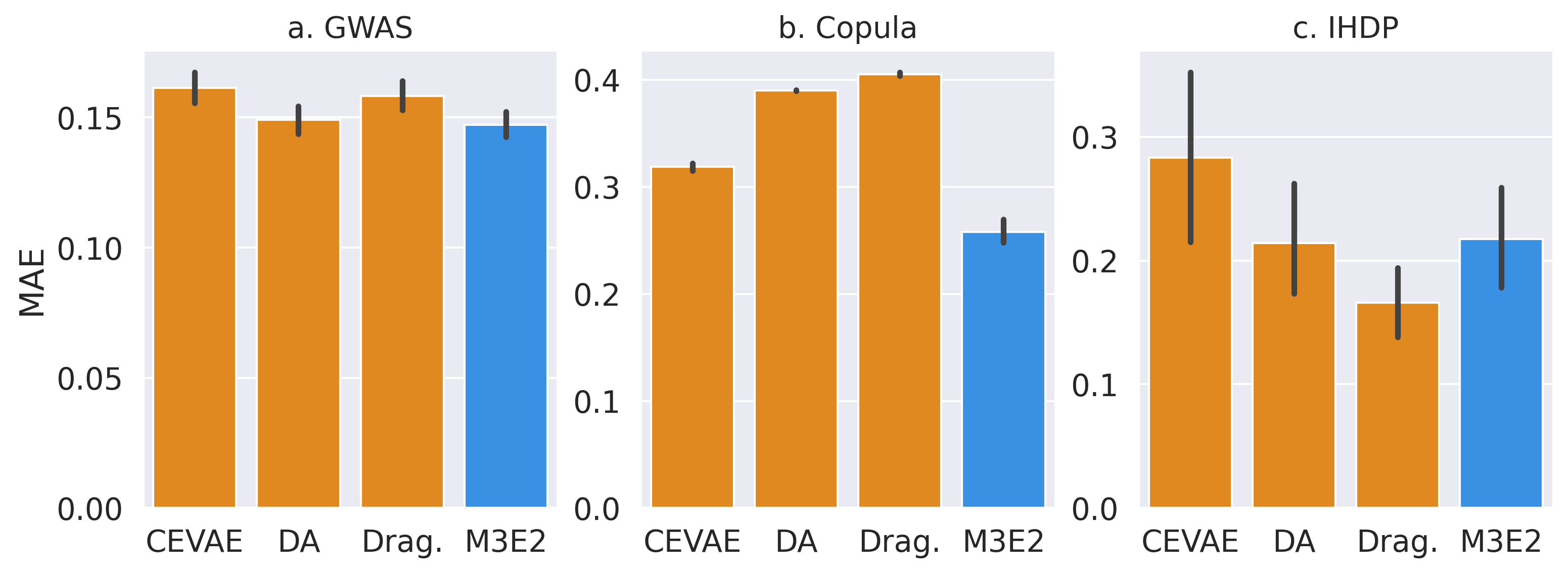}}
    \caption{ MAE barplots of the M3E2 and baseline methods. Small MAE values are desirable. The black line indicates a 95\% confidence interval.}
    \label{ex1}
\end{figure}

Figure \ref{ex1} shows, for each dataset, the average MAE across all settings. Our proposed method, M3E2, clearly outperforms all baselines on the multi-treatment datasets GWAS and COPULA. On IHDP, a single-treatment dataset, M3E2 was outperformed by Dragonnet, yet, it was better than the other two baselines. Note that our results for Dragonnet on IHDP match the results previously reported \cite{shi2019adapting}, and the estimators' larger variance on the IHDP dataset can be explained by the scale of the true treatment effect. Our main take from Figure \ref{ex1} is that our method outperforms all the baselines on its ideal use-case: applications with multiple treatment effects. In single-treatment applications, while achieving reasonable results, simpler architectures that target single-treatment estimation like the Dragonnet tend to achieve better performance.

 %M3E2 has the best performance with the lowest MAE among the methods considered on both benchmark datasets.
 \begin{figure}[!h]
    \centerline{\includegraphics[width=0.7\textwidth, scale=0.45]{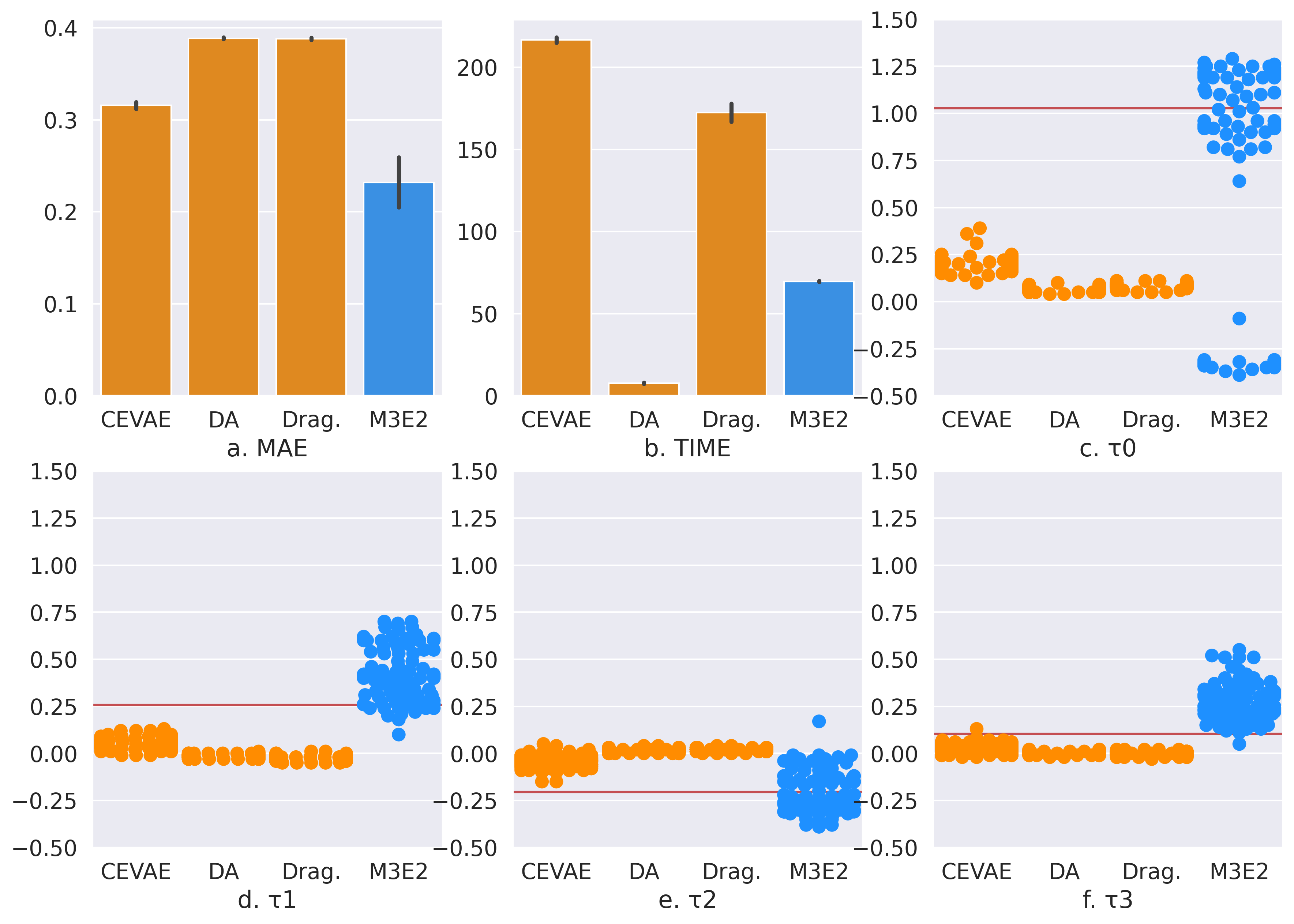}}
\caption{Copula results for one simulated dataset ($n=10000, k=4, v=10$) with 24 independent repetitions of each model. The baselines' results are shown in orange, our results are in blue, and the red line shows the true effect (c-f).}
    \label{ex2}
\end{figure}

Figure \ref{ex2} shows a deeper analysis of the Copula dataset. Figure \ref{ex2}.a shows that M3E2 has the lowest MAE values compared to the other baselines. Figure \ref{ex2}.b shows the total run time of each method in seconds. As a reminder, both DA and M3E2 fit one model for all treatments; Dragonnet and CEVAE, on the other hand, fit one model for each treatment. DA, a probabilistic model, has the fastest running time; M3E2 has the lowest running time among the NN methods. A comparison between the true $\tau$ (line in red) and the estimated treatment effects (dots) is shown in Figures \ref{ex2}.c-f. Note that for $\tau_0$ and $\tau_2$, M3E2 is the only method whose estimates are centered around the true value. For $\tau_1$ and $\tau_3$, M3E2 overestimates the treatment effects, yet, it still produces reasonably good estimates. Overall, M3E2 has a good performance. However, we noticed two limitations: First, M3E2 has a larger variance than the other methods; second, for some runs, it estimated values very far from the true treatment effect $\tau_0$. Considering our baselines, while they have a smaller variance, we noticed that DA and Dragonnet often estimated the treatment effect as 0, indicating that these methods might fail to estimate the treatment effect in this dataset correctly, despite achieving reasonable predictive performance. CEVAE was the second-best method; still, its results were never centered around the true values (red lines) and often underestimated the magnitude of the treatment effect.

\subsection{Impact of Dataset Parameters}

\begin{figure}[!h]
    \centerline{\includegraphics[width=0.8\textwidth]{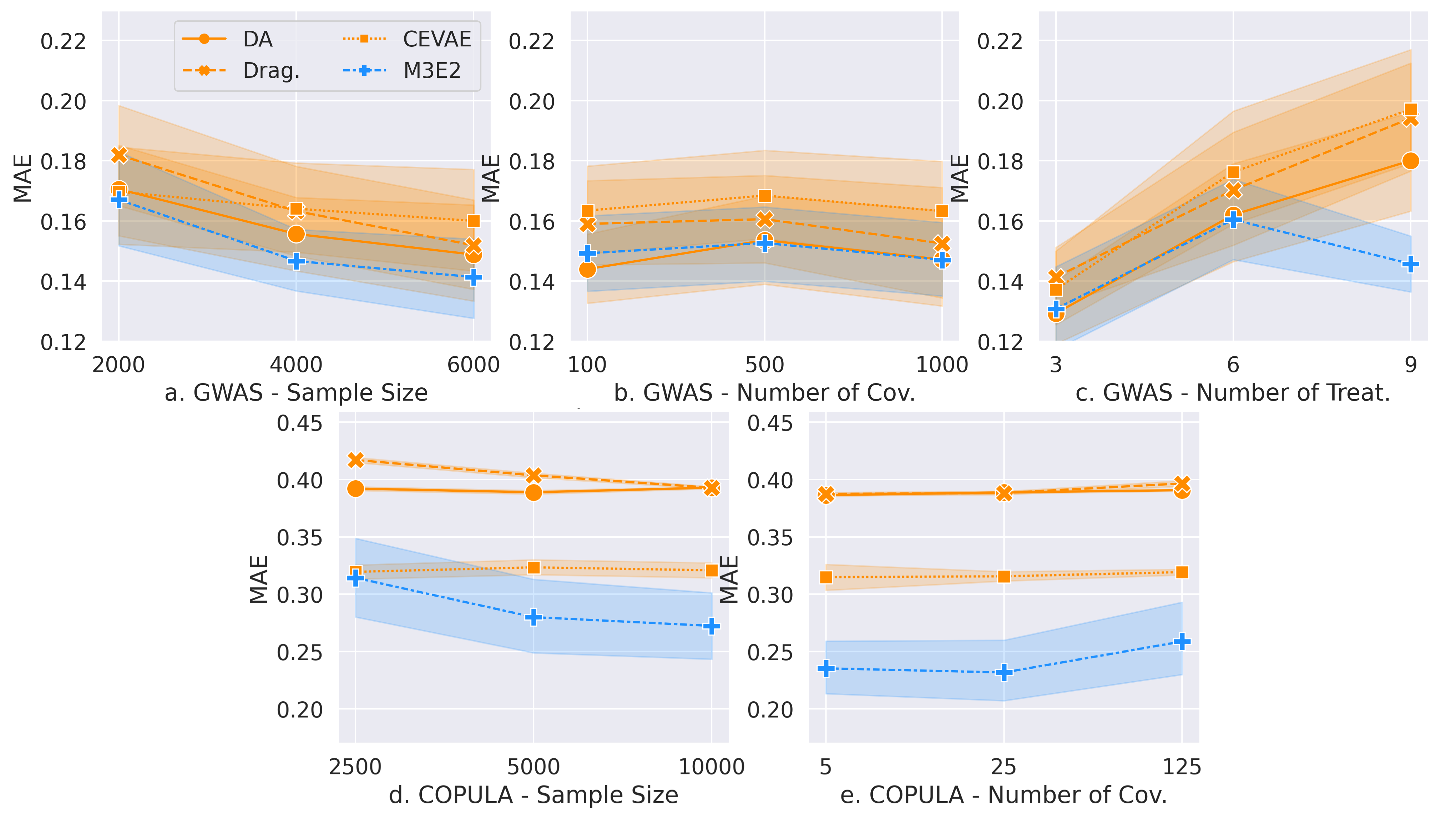}}
    \caption{Impact of the dataset parameters in estimating multiple treatment effects.}
    \label{plot_lines}
\end{figure}

We also explored the impact of the dataset parameters in estimating the multiple treatment effects. We focused on three parameters: the sample size, number of treatments, and covariates. Figure \ref{plot_lines} shows, in detail, the average MAE and the 95\% confidence interval (colored area) for the several settings. Figure \ref{plot_lines}.$a$ and \ref{plot_lines}.$d$ show the impact of the sample size on the GWAS and Copula dataset, respectively. Our proposed method, M3E2, is the method that benefits the most from increasing the sample size. We noticed that all methods are robust to the increase in the number of covariates (Figures  \ref{plot_lines}.b and \ref{plot_lines}.e), with M3E2 having a small increase on MAE on the Copula dataset with 125 covariates. The most surprising result of all is shown in Figure \ref{plot_lines}.c. The MAE increases in all baselines with the increase in the number of treatments. Nevertheless, M3E2 achieves better results with nine treatments than with six treatments. Such a result shows that, while the methods are similar regarding the dataset impact on MAE and are quite robust to variations in the number of covariates, M3E2 significantly outperforms all other methods when a larger number of treatment effects are considered.

\section{Discussion and Conclusion}\label{discussion}

In this paper, we have investigated the problem of estimating the effect of multiple treatments in observational data, a setting often found in biomedical applications. To address current limitations, we proposed the M3E2, a multiple treatment effect estimator that uses a MTL neural network architecture. One of the main advantages of M3E2 is its flexibility, as several of its subcomponents can be replaced by alternative implementations, e.g., by different experts, latent variable models, or propensity score predictors. We experimentally compared M3E2 against three baselines on three synthetic benchmark datasets that mimic biomedical applications. The online repository \url{github.com/raquelaoki/M3E2} contains the code to replicate all the experiments, and we put extra effort into making the M3E2 implementation agnostic to the application; therefore, its deployment in other applications should be straightforward. M3E2 demonstrated promising experimental results and strong evidence that MTL contributed to more accurate estimates of the treatment effects. Nevertheless, there remain several directions for future research. As discussed in Section \ref{assumption}, our method assumes ignorability, which is quite limiting in real-life applications. M3E2 also inherits the limitations of other MTL models, in particular, the susceptibility to imbalanced tasks and overfitting. All strengths and limitations considered, we believe that M3E2 has a very good use case with manageable limitations. In future research, we want to apply our proposed method to a real-world dataset that records adverse drug reactions in therapies for treating cancer in infants, moving a step forward toward the precision medicine goal of providing the \textit{right drug at the right dose to the right patient}\cite{collins2015new}.

\textbf{Funding Statement:} This project is supported by Genome Canada, Genome British Columbia (funding $\#$272PGX) and the Canadian Institutes of Health Research (funding $\#$GP1-155872) through the Large Scale Applied Research Project Competition. Additional support has been provided as required by the grant competition by British Columbia's Provincial Health Services Authority (PHSA); BC Children's Hospital Foundation; Health Canada (in-kind); Illumina (in-kind) and Thermo Fisher Scientific (Canada).

\bibliographystyle{ws-procs11x85}
\bibliography{references}

\end{document}